\documentclass[sn-mathphys,Numbered]{sn-jnl}

\usepackage{xcolor}%
\definecolor{LColor}{HTML}{7cd2f7} 
\definecolor{RColor}{HTML}{f7c265} 
\definecolor{MColor}{HTML}{EDF4F4} 
\definecolor{BColor}{HTML}{F7F7F7} 
\usepackage{pgfplots}%
\pgfplotsset{compat=1.17}

\usepackage{graphicx}%
\usepackage{multirow}%
\usepackage{amsmath,amssymb,amsfonts}%
\usepackage{amsthm}%
\usepackage{mathrsfs}%
\usepackage[title]{appendix}%
\usepackage{xcolor}%
\usepackage{textcomp}%
\usepackage{manyfoot}%
\usepackage{booktabs}%
\usepackage{algorithm}%
\usepackage{algorithmicx}%
\usepackage{algpseudocode}%
\usepackage{listings}%
\usepackage{caption}%
\usepackage[T1]{fontenc}
\usepackage{listings}
\usepackage{color}
\usepackage{array}
\usepackage{booktabs} 
\usepackage{geometry}  

\definecolor{codegreen}{rgb}{0,0.6,0}
\definecolor{codegray}{rgb}{0.5,0.5,0.5}
\definecolor{codepurple}{rgb}{0.58,0,0.82}
\definecolor{backcolour}{rgb}{0.95,0.95,0.92}

\lstdefinestyle{jsonstyle}{
    backgroundcolor=\color{backcolour},
    commentstyle=\color{codegray},
    keywordstyle=\color{magenta},
    numberstyle=\tiny\color{codegray},
    stringstyle=\color{codepurple},
    basicstyle=\ttfamily\small,
    breakatwhitespace=false,
    breaklines=true,
    captionpos=b,
    keepspaces=true,
    numbers=none,
    showspaces=false,
    showstringspaces=false,
    showtabs=false,
    tabsize=2
}

\theoremstyle{thmstyleone}%
\theoremstyle{thmstyletwo}%

\theoremstyle{thmstylethree}%

\begin{document}

\title[Article Title]{Integrating UMLS Knowledge into Large Language Models for Medical Question Answering}


\author*[1,2]{\fnm{Rui} \sur{Yang}}\email{yang\_rui@u.nus.edu}

\author[3,4]{\fnm{Edison} \sur{Marrese-Taylor}}

\author[5]{\fnm{Yuhe} \sur{Ke}}

\author[6]{\fnm{Lechao} \sur{Cheng}}

\author[7.8]{\fnm{Qingyu} \sur{Chen}}

\author[3]{\fnm{Irene} \sur{Li}}

\affil*[1]{\orgname{National University of Singapore}}

\affil[2]{\orgname{Duke-NUS Medical School}}

\affil[3]{\orgname{University of Tokyo}}

\affil[4]{\orgname{National Institute of Advanced Industrial Science and Technology}}

\affil[5]{\orgname{Singapore General Hospital}}

\affil[6]{\orgname{Zhejiang Lab}}

\affil[7]{\orgname{National Institutes of Health}}

\affil[8]{\orgname{Yale University}}



\abstract{Large language models (LLMs) have demonstrated powerful text generation capabilities, bringing unprecedented innovation to the healthcare field. While LLMs hold immense promise for applications in healthcare, applying them to real clinical scenarios presents significant challenges, as these models may generate content that deviates from established medical facts and even exhibit potential biases. In our research, we develop an augmented LLM framework based on the Unified Medical Language System (UMLS), aiming to better serve the healthcare community. We employ LLaMa2-13b-chat and ChatGPT-3.5 as our benchmark models, and conduct automatic evaluations using the ROUGE Score and BERTScore on 104 questions from the LiveQA test set. Additionally, we establish criteria for physician evaluation based on four dimensions: Factuality, Completeness, Readability and Relevancy. ChatGPT-3.5 is used for physician evaluation with 20 questions on the LiveQA test set. Multiple resident physicians conducted blind reviews to evaluate the generated content, and the results indicate that this framework effectively enhances the factuality, completeness, and relevance of generated content. Our research demonstrates the effectiveness of using UMLS-augmented LLMs and highlights the potential application value of LLMs in medical question-answering.}

\maketitle

\section{Introduction}\label{sec1}

Large language models (LLMs) such as GPT-4 \cite{openai2023gpt4} and LLaMa2 \cite{touvron2023llama} have demonstrated powerful text generation capabilities and are bringing about transformative changes across various domains. Similarly, LLMs have ample potential for application in healthcare \cite{yang2023large, li2022ehrkit, li2022neural}. However, a critical concern arises from the potential of these models to generate content that deviates from established medical facts. This discrepancy primarily stems from the models' inadequacy in pre-training on medical data, making them unable to discern the veracity of medical information \cite{xie2023faithful}. Moreover, LLM-generated content can perpetuate and amplify biases related to ethnicity, gender, and socioeconomic background \cite{yang2023large}. Despite the broad research prospects of LLMs in healthcare, these issues impede the further application of LLMs in real clinical settings.


The existing medical LLMs typically use medical data for optimization during the fine-tuning or prompt-tuning process to enhance their medical knowledge, such as Med-PaLM2 \cite{singhal2023expertlevel}, baize-healthcare \cite{xu2023baize}, and more. However, this process relies on computationally intensive resources and high-quality data, limiting the model's acquired medical knowledge to the specific data utilized for tuning. Furthermore, fine-tuned or prompt-tuned medical LLMs face challenges in keeping up with the constantly evolving medical knowledge, as they cannot update their knowledge in real-time.

Other related studies have highlighted that leveraging external medical knowledge bases can help alleviate these issues. Almanac \cite{hiesinger2023almanac} is an LLM framework for clinical medicine that can generate more accurate and reliable content by retrieving information from the Internet. ChatENT \cite{long2023chatent} is an LLM specifically designed for Otolaryngology - Head and Neck Surgery (OHNS). The LLM is augmented by retrieving relevant information from the OHNS database, which is constructed using information accessed from the internet. However, retrieving external knowledge, especially from the Internet, to answer medical questions might introduce more noise. The irrelevant information might in turn compromise the effectiveness of the LLM \cite{malaviya2023expertqa}. Concurrently, there are concerns about information credibility, data consistency, potential privacy and security risks, as well as copyright and legal issues.


In this study, we propose integrating a structured medical knowledge graph into existing pretrained LLMs to achieve trustable and explainable medical question-answering (QA) with supporting evidence. By doing so, we are able to use the well-structured domain knowledge to guide LLMs in generating more factual answers. Our framework retrieves definitions and relations of relevant medical terminologies from The Unified Medical Language System (UMLS) database and prompts to a given LLM. We select LLaMa2-13b-chat and ChatGPT-3.5 as our benchmark models and conduct automatic evaluations. In addition, to provide a more reliable assessment of the generated content, we establish an evaluation criteria based on four dimensions: Factuality, Completeness, Readability and Relevancy. We further employ ChatGPT-3.5 for the human evaluation, and the assessment was conducted by multiple physicians in a blind review process following the predefined criteria. The results reveal significant concordance, highlighting the potential of our approach. 

\section{Method}\label{sec2}

\begin{figure}[h]
\centering
\includegraphics[width=1\linewidth]{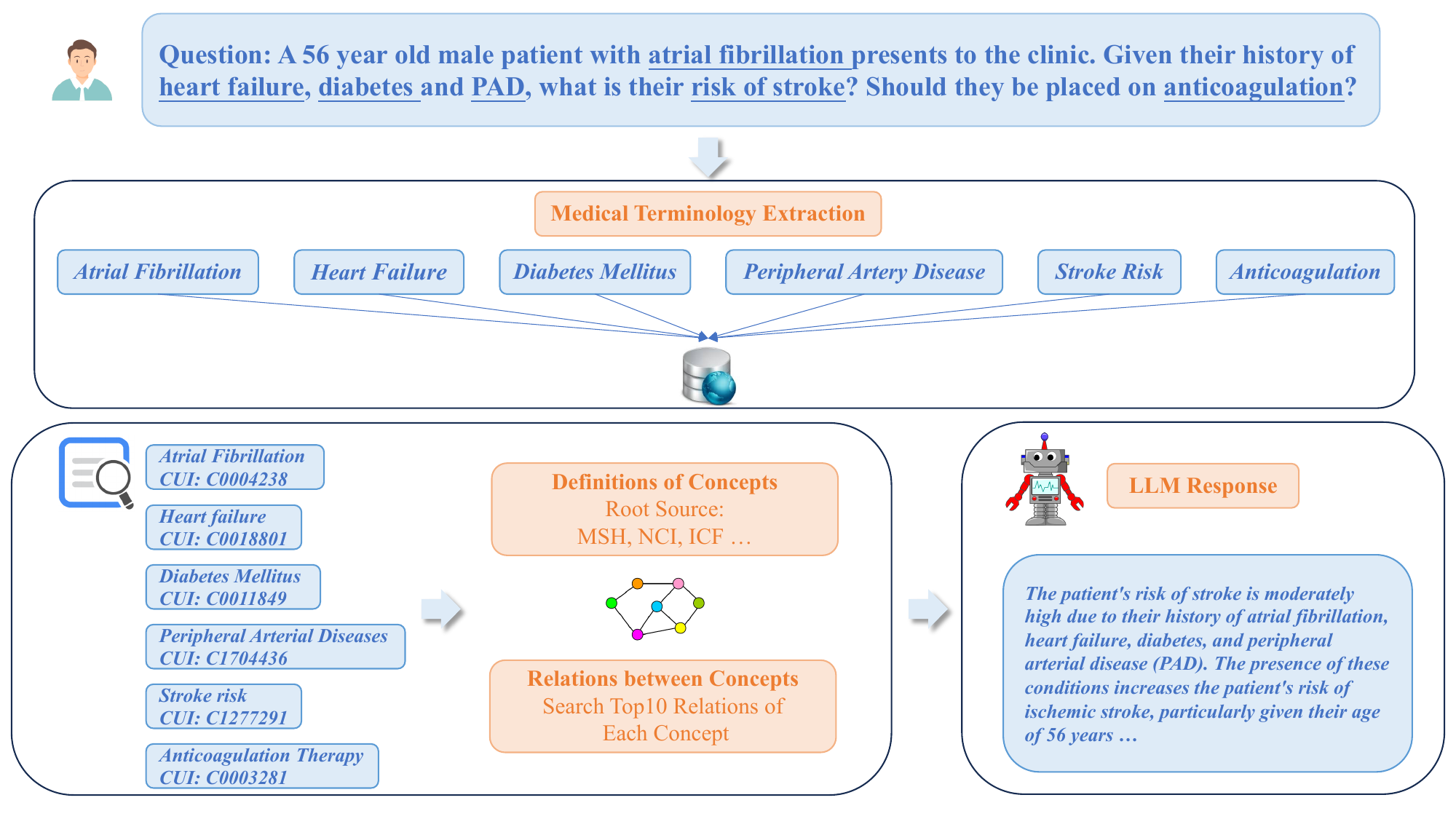}
\caption{Framework for Augmenting LLMs with UMLS Database.}
\label{fig:framework}
\end{figure}

In this research, we enhance the medical knowledge of LLMs by leveraging UMLS database. The specific framework is shown in Fig.~\ref{fig:framework}. The input is a medical question proposed by the user. After receiving the input, we first extract medical terminologies from the given question text, such as ``Atrial Fibrillation", ``Heart Failure" and so on. High-quality extraction of medical terminology is crucial, and we discuss the strengths and weaknesses of three different extraction techniques in the subsequent sections. In the second step, we map these medical terminologies to their corresponding concepts and get their Concept Unique Identifiers (CUIs) using the UMLS API. Then, we extract the definitions corresponding to the concepts and their relations with other medical terminologies through the CUIs. We prompt the LLM with this additional medical information to help it respond to the question. Finally, LLM will provide the user with the answer and offer trustworthy explanations.

\subsection{UMLS}\label{subsec1}

UMLS \cite{bodenreider2004unified}, developed by The United States National Library of Medicine (NLM), is a repository of health and biomedical vocabularies. It provides a comprehensive, standardized list of medical terminology, enabling data extraction in a structured and factual manner. UMLS consists of three primary knowledge sources: the Metathesaurus, the Semantic Network, and the SPECIALIST Lexicon and Lexical Tools. Metathesaurus as a core component of UMLS, which includes a collection of 14 million names sourced from 209 different terminologies and support for 25 languages. Metathesaurus connects names from different source terminologies that may have different forms and codes but share the same or similar concepts, and maps them to the same CUI. For example, (``Addison Disease'', ``Mesh'', ``D000224''), (``Addison's disease'', ``ICD-10-CM'', ``E27.1''), (``Primary adrenocortical insufficiency (disorder)'', ``SNOMED CT'', ``373662000''), (``Primary hypoadrenalism'', ``MedDRA'', ``S0718109''), (``Enfermedad de Addison'', ``Mesh Spanish'', ``D000224''), and so on, can all be mapped to the same CUI: C0001403. And within Metathesaurus, there are more than 3.8 million concepts and 78 million relationships, which can provide extensive coverage of medical knowledge for LLMs. Previous research has shown that integrating UMLS into Pre-trained Models (PLMs) can help models take advantage of domain-specific knowledge, improving their performance on medical tasks \cite{huang2020biomedical,he2019integrating,lu2021parameter,aracena2022knowledge,gao2023leveraging}. 

\subsection{Medical Terminology Extraction}\label{subsec2}

Medical terminology extraction is a highly important step. The LLM will utilize the extracted medical terms to search the UMLS database for their definitions and relations. This UMLS knowledge will effectively assist the LLM in answering medical questions. Meanwhile, incorrect extraction or missing extraction will prevent the model from obtaining accurate and sufficient medical knowledge.

In light of this important issue, in this paper we meticulously assess three alternative techniques for extracting medical terminologies, including Instruction LLM (Direct Extraction), Instruction LLM (Indirect Extraction), and Biomedical PLM for Named Entity Recognition (NER). The first two techniques utilize designed prompts to enable LLM to extract medical terminologies automatically, as shown in Tab.~\ref{tab:designed prompt}. Notably, direct extraction approach returns medical terminologies explicitly appearing in the question text, whereas indirect extraction method extracts terminologies that, while contextually relevant, may not manifest verbatim in the question text. As for NER technique, we utilize the most downloaded Bio-Epidemiology-NER \cite{raza2022large} model from HuggingFace, which can identify 107 types of entities within the biomedical domain.

\begin{table}[h]
    \centering
    \begin{tabular}{|m{0.9\linewidth}|}  
        \hline
        \multicolumn{1}{|c|}{\textbf{Direct Extraction}} \\
        \hline
        Only return the medical terminologies contained in the input question. \\
        Please return in JSON format. \\
        \textbf{Output Format:}
        \begin{lstlisting}
{
  "medical terminologies": ["<name>", "<name>"]
}
        \end{lstlisting}
        Please only return the JSON format information. \\
        \textbf{Input:} \{question\} \\
        \textbf{Output:} \\
        \hline
        \multicolumn{1}{|c|}{\textbf{Indirect Extraction}} \\
        \hline
        Return medical terminologies related to the input question. \\
        Please return in JSON format. \\
        \textbf{Output Format:}
        \begin{lstlisting}
{
  "medical terminologies": ["<name>", "<name>"]
}
        \end{lstlisting}
        Please only return the JSON format information. \\
        \textbf{Input:} \{question\} \\
        \textbf{Output:} \\
        \hline
    \end{tabular}
\caption{Designed Prompts for Direct Extraction and Indirect Extraction.}
\label{tab:designed prompt}
\end{table}

\begin{figure}[h]
\centering
\includegraphics[width=1\linewidth]{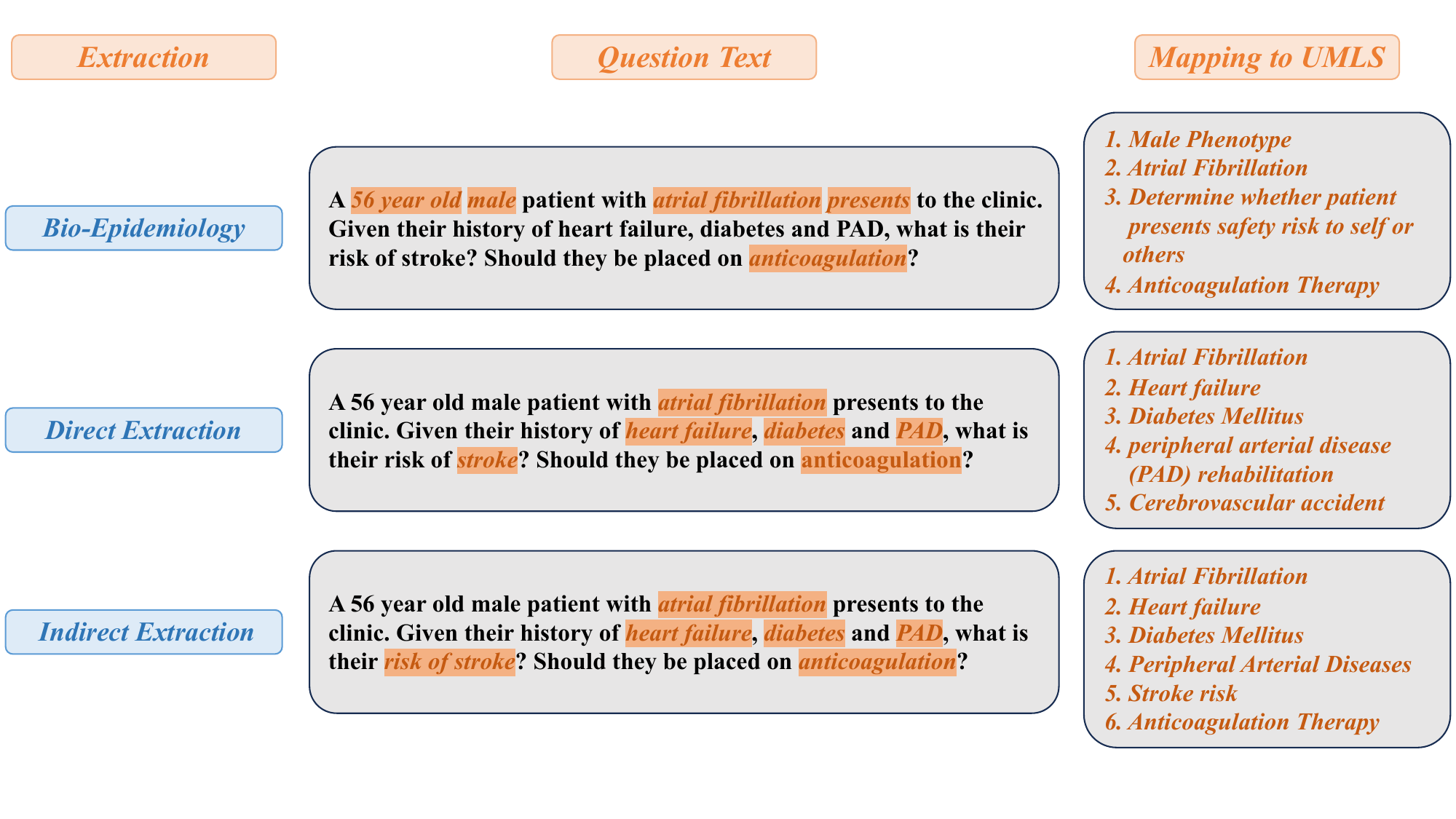}
\caption{Medical Terminology Extraction Techniques: Bio-Epidemiology-NER, Direct Extraction (LLM: LLaMa2-13b-chat), Indirect Extraction (LLM: LLaMa2-13b-chat) and Corresponding UMLS Mappings.}
\label{fig:extraction}
\end{figure}

Fig~\ref{fig:extraction} shows an example of medical terminology identification utilizing the aforementioned three techniques. As can be observed, the Bio-Epidemiology-NER model fails to recognize several important medical concepts present in the question, such as ``heart failure'', ``diabetes'', ``PAD'', and ``stroke''. Conversely, it discerns entities that do not require a UMLS search, including gender and age. Compared to these results, we see that two methods of Instruction LLM identify more relevant entities. In addition, through more testing, Instruction LLM outperforms the Biomedical NER model. However, direct extraction and indirect extraction each have their pros and cons. When multiple similar medical terminologies appear in a single question, direct extraction repeats the extraction, increasing the UMLS retrieval steps and slowing down LLM's inference speed. On the other hand, indirect extraction does not repeatedly extract similar medical terminologies but may introduce many new terms. The introduction of extraneous terms can lead LLM to retrieve irrelevant content from UMLS, which cannot prompt LLM to answer questions accurately. Furthermore, direct extraction lacks flexibility, whereas indirect extraction demonstrates more significant flexibility. For instance, it automatically recognizes ``PAD'' to "Peripheral Artery Disease" and extracts ``risk of stroke'' as ``Stroke Risk'' instead of merely returning ``stroke''. This distinction holds significant relevance in UMLS, as querying ``stroke'' will return the concept ``Cerebrovascular accident'', whereas querying ``Stroke Risk'' will return the concept ``Stroke risk''. Clearly, the latter is more in line with our desired outcome. Both of these techniques will be used in the subsequent evaluations.

\subsection{Definition \& Relation Extraction}\label{subsec3}

We believe that medical information from multiple reliable sources will significantly enrich LLM's knowledge base, and the definitions of medical terms contained in the question and their relations with other terms will help LLM better understand the question and make inference \cite{li2021unsupervised}.

To that end, we proceed to extract definitions and relation information using the CUI. Each CUI may return multiple definitions from various sources, including ``Medical Subject Headings (MSH)'', ``National Cancer Institute (NCI)'', ``International Classification of Functioning, Disability and Health (ICF)'', and others. We prioritize English language definition and return it. For a given CUI, there may be dozens or even hundreds of relations. Considering computational memory and the inference speed of the LLM, we extract only the top 25 relations. These definitions and relationship information will be returned and input into the LLM prompt to assist the model in generating answers.

\subsection{Data \& Evaluation Criteria}\label{subsec4}

In our research, we conduct evaluations using the LiveQA dataset \cite{abacha2017overview} from Text Retrieval Challenge (TREC) 2017, which consists of health questions submitted by consumers to the National Library of Medicine (NLM). The LiveQA dataset comprises a training set with 634 question-answer pairs and a test set with 104 pairs. The test set was utilized for automatic scoring evaluation and physician evaluation. For the automated evaluation, we employ ChatGPT-3.5 and LLaMa2-13b-chat as our benchmark models, utilizing both the ROUGE Score \cite{lin2004rouge} and BERTScore \cite{zhang2019bertscore} for assessment. However, these metrics have been shown to be inaccurate in assessing the quality of generated content, and the reference answers lack revision by medical experts to serve as ground truth. To ensure reliability and fairness, we establish criteria for physician-based evaluation using four dimensions: Factuality, Completeness, Readability, and Relevancy, as shown in Tab.~\ref{tab:evaluation_criteria}. A total of 20 questions are extracted from the test set for evaluation. Multiple physicians conducted a blind evaluation of the answers generated by ChatGPT-3.5 and our UMLS-augmented version, and the results were highly consistent. In the next step, we will include more questions for physician-evaluation to further ensure the reliability of the results.


\begin{table}[h]
\centering
\renewcommand{\arraystretch}{1.5} 
\setlength{\tabcolsep}{12pt} %
\begin{tabular}{c p{9.8cm}}
\toprule 
 & \multicolumn{1}{c}{\textbf{Description}} \\
\midrule 
\textbf{Factuality} & The degree to which the generated text aligns with established medical facts, providing accurate explanations for further verification. \\
\textbf{Completeness} & The degree to which the generated text comprehensively portrays the clinical scenario or posed question, including other pertinent considerations. \\
\textbf{Readability} & The extent to which the generated text is readily comprehensible to the user, incorporating suitable language and structure to facilitate accessibility. \\
\textbf{Relevance} & The extent to which the generated text directly addresses medical questions while encompassing a comprehensive range of pertinent information. \\
\bottomrule 
\end{tabular}
\caption{Evaluation Criteria for Generated Content}
\label{tab:evaluation_criteria}
\end{table}

\section{Results}\label{sec3}

The automated scoring results are shown in Tab.~\ref{table:evaluation}. In the first group, it can be observed that ChatGPT-3.5 achieved higher scores in both ROUGE Score and BERTScore. While in the second group, UMLS-augmented LLaMa2-13b-chat performed better in most of the scores. However, as previously mentioned, the automated scoring results lack reliability, whereas physician evaluation offers a more dependable assessment.

\begin{table}[h]
\centering
\begin{tabular}{ccccccc}
\toprule
 & \multicolumn{3}{c}{ROUGE} & \multicolumn{3}{c}{BERTScore} \\
 & R-1 & R-2 & R-L & P & R & F1 \\
\midrule
ChatGPT-3.5 & \textbf{21.44} & \textbf{4.44} & \textbf{19.65} & \textbf{83.15} & \textbf{83.21} & \textbf{83.14} \\
ChatGPT-3.5$_{\text{*D.E.+UMLS}}$ & 21.33 & 4.23 & 19.61 & 82.80 & 82.85 & 82.79 \\
ChatGPT-3.5$_{\text{*I.E.+UMLS}}$ & 20.85 & 3.94 & 19.17 & 82.72 & 82.73 & 82.69 \\
\midrule
LLaMa2-13b-chat & 19.07 & 3.51 & 17.68 & 80.83 & \textbf{83.07} & 81.91 \\
LLaMa2-13b-chat$_{\text{*D.E.+UMLS}}$ & \textbf{19.97} & \textbf{3.58} & \textbf{18.24} & 81.50 & 82.74 & 82.08 \\
LLaMa2-13b-chat$_{\text{*I.E.+UMLS}}$ & 19.66 & 3.44 & 18.10 & \textbf{81.53} & 82.78 & \textbf{82.12} \\
\bottomrule
\end{tabular}
\caption{ROUGE Score and BERTScore on Different Settings. D.E.: Direct Extraction for Medical Terminologies. I.E.: Indirect Extraction for Medical Terminologies.}
\label{table:evaluation}
\end{table}

\begin{figure}[h]
\centering
\begin{tikzpicture}[scale=1]
    \begin{axis}[
            axis background/.style={fill=BColor!70},
            xbar stacked,
            width=11cm, height=5.5cm,
            nodes near coords={\pgfmathprintnumber\pgfplotspointmeta\%},
            xmin=0, xmax=102,
            y axis line style = {BColor},
            x axis line style = {BColor},
            axis y line=left, 
            xlabel={Percentage (\%)},
            bar width=15pt,
            xtick style={draw=none},
            ytick style={draw=none}, 
            enlarge y limits=0.2, 
            legend style={at={(0.5,1.20)}, 
            anchor=north, legend columns=-1},
            symbolic y coords={Better Completeness, Better Readability, Better Relevancy, Better Factuality},
            xtick={0, 20, 40, 60, 80, 100},
        ]
        \addplot[draw=none,fill=LColor] coordinates {(55,Better Completeness) (40,Better Readability) (25,Better Relevancy) (40,Better Factuality)};
        \addplot[draw=none,fill=MColor] coordinates {(25,Better Completeness) (15,Better Readability) (60,Better Relevancy) (30,Better Factuality)};
        \addplot[draw=none,fill=RColor] coordinates {(20,Better Completeness) (45,Better Readability) (15,Better Relevancy) (30,Better Factuality)};
        
        \legend{\strut UMLS-augmented ChatGPT-3.5,\! \strut Tie,\! \strut ChatGPT-3.5}
    \end{axis}
\end{tikzpicture}
\caption{Win rate of content generated by ChatGPT-3.5 and our UMLS-augmented version, as evaluated by physicians.}
\label{fig:physician-evaluation}
\end{figure}
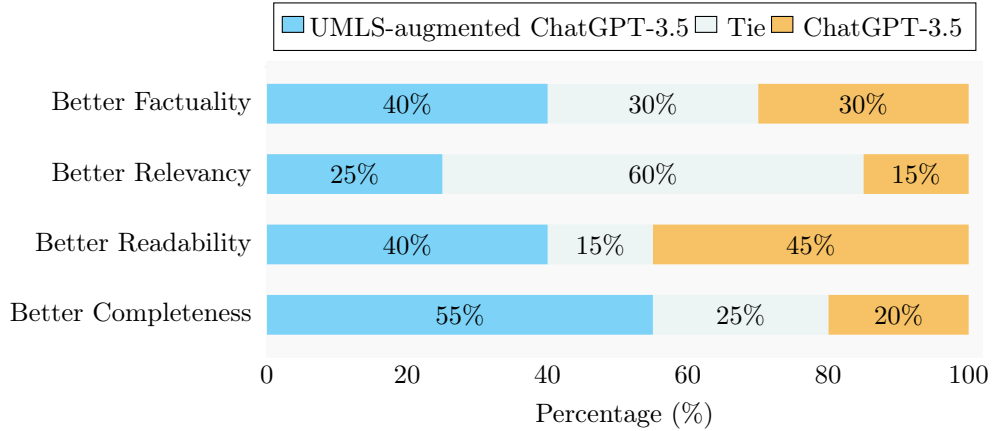

The results of physician evaluation are shown in Fig.~\ref{fig:physician-evaluation}. In terms of Factuality, both UMLS-augmented ChatGPT-3.5 and ChatGPT-3.5 demonstrated comparable performance in 30\% of the evaluated questions. For 40\% of these questions, physicians believed that UMLS-augmented ChatGPT-3.5 exhibited superior performance relative to ChatGPT-3.5, while ChatGPT-3.5 surpassed UMLS-augmented ChatGPT-3.5 in performance for another 30\% of the questions. In terms of Relevance, the performance of both was on par for 60\% of the questions, but UMLS-augmented ChatGPT-3.5 won more of the remaining questions, surpassing ChatGPT-3.5 by 10\%. As for Readability, ChatGPT-3.5 secured a win rate of 45\%, in contrast to the 40\% rate of UMLS-augmented ChatGPT-3.5. This could be due to the more specialized medical information provided by UMLS, which reduces the readability for users without a medical background. From a perspective of content Completeness, UMLS-augmented ChatGPT-3.5 notably outperformed ChatGPT-3.5, securing a win rate of 55\%.

\section{Limitations}\label{sec4}
 
In this research, we propose a LLM framework augmented by UMLS to improve the quality of the content generated, especially in terms of factuality. However, there are some limitations, which we will address in the next phase.

\textbf{UMLS Information Extraction}: When retrieving relation information of medical terminologies from UMLS database, we automatically extract the top 25 relations. However, these 25 relations may not directly pertain to the specific question, and irrelevant information cannot aid the LLM in generating high-quality content. Moreover, truly relevant relations may not be extracted, rendering the answers incomplete. In the next step, we will extract all relations of medical terminologies and compute the most relevant relation information to prompt LLM.

\textbf{Evaluation}: We employ both automated scoring and physician-evaluation as our assessment methods. However, the LiveQA dataset we used lacks a ground truth verified by medical experts. Moreover, automatic evaluation metrics such as ROUGE Score and BERTScore may not accurately reflect the quality of the generated content. Although we conduct a human study with physicians, it was limited to 20 questions, which we think may set clear boundaries to our assessment results. To further enhance the reliability of our evaluation, in the next step, we will use dataset revised by medical experts for automated assessments. Additionally, we will involve more physicians to conduct in-depth manual evaluations across a broader range of questions.


\bibliography{sn-bibliography}


\begin{thebibliography}{22}
\ifx \bisbn   \undefined \def \bisbn  #1{ISBN #1}\fi
\ifx \binits  \undefined \def \binits#1{#1}\fi
\ifx \bauthor  \undefined \def \bauthor#1{#1}\fi
\ifx \batitle  \undefined \def \batitle#1{#1}\fi
\ifx \bjtitle  \undefined \def \bjtitle#1{#1}\fi
\ifx \bvolume  \undefined \def \bvolume#1{\textbf{#1}}\fi
\ifx \byear  \undefined \def \byear#1{#1}\fi
\ifx \bissue  \undefined \def \bissue#1{#1}\fi
\ifx \bfpage  \undefined \def \bfpage#1{#1}\fi
\ifx \blpage  \undefined \def \blpage #1{#1}\fi
\ifx \burl  \undefined \def \burl#1{\textsf{#1}}\fi
\ifx \doiurl  \undefined \def \doiurl#1{\url{https://doi.org/#1}}\fi
\ifx \betal  \undefined \def \betal{\textit{et al.}}\fi
\ifx \binstitute  \undefined \def \binstitute#1{#1}\fi
\ifx \binstitutionaled  \undefined \def \binstitutionaled#1{#1}\fi
\ifx \bctitle  \undefined \def \bctitle#1{#1}\fi
\ifx \beditor  \undefined \def \beditor#1{#1}\fi
\ifx \bpublisher  \undefined \def \bpublisher#1{#1}\fi
\ifx \bbtitle  \undefined \def \bbtitle#1{#1}\fi
\ifx \bedition  \undefined \def \bedition#1{#1}\fi
\ifx \bseriesno  \undefined \def \bseriesno#1{#1}\fi
\ifx \blocation  \undefined \def \blocation#1{#1}\fi
\ifx \bsertitle  \undefined \def \bsertitle#1{#1}\fi
\ifx \bsnm \undefined \def \bsnm#1{#1}\fi
\ifx \bsuffix \undefined \def \bsuffix#1{#1}\fi
\ifx \bparticle \undefined \def \bparticle#1{#1}\fi
\ifx \barticle \undefined \def \barticle#1{#1}\fi
\bibcommenthead
\ifx \bconfdate \undefined \def \bconfdate #1{#1}\fi
\ifx \botherref \undefined \def \botherref #1{#1}\fi
\ifx \url \undefined \def \url#1{\textsf{#1}}\fi
\ifx \bchapter \undefined \def \bchapter#1{#1}\fi
\ifx \bbook \undefined \def \bbook#1{#1}\fi
\ifx \bcomment \undefined \def \bcomment#1{#1}\fi
\ifx \oauthor \undefined \def \oauthor#1{#1}\fi
\ifx \citeauthoryear \undefined \def \citeauthoryear#1{#1}\fi
\ifx \endbibitem  \undefined \def \endbibitem {}\fi
\ifx \bconflocation  \undefined \def \bconflocation#1{#1}\fi
\ifx \arxivurl  \undefined \def \arxivurl#1{\textsf{#1}}\fi
\csname PreBibitemsHook\endcsname

\bibitem[\protect\citeauthoryear{OpenAI}{2023}]{openai2023gpt4}
\begin{botherref}
\oauthor{\bsnm{OpenAI}}:
GPT-4 Technical Report
(2023)
\end{botherref}
\endbibitem

\bibitem[\protect\citeauthoryear{Touvron et~al.}{2023}]{touvron2023llama}
\begin{botherref}
\oauthor{\bsnm{Touvron}, \binits{H.}},
\oauthor{\bsnm{Martin}, \binits{L.}},
\oauthor{\bsnm{Stone}, \binits{K.}},
\oauthor{\bsnm{Albert}, \binits{P.}},
\oauthor{\bsnm{Almahairi}, \binits{A.}},
\oauthor{\bsnm{Babaei}, \binits{Y.}},
\oauthor{\bsnm{Bashlykov}, \binits{N.}},
\oauthor{\bsnm{Batra}, \binits{S.}},
\oauthor{\bsnm{Bhargava}, \binits{P.}},
\oauthor{\bsnm{Bhosale}, \binits{S.}}, et al.:
Llama 2: Open foundation and fine-tuned chat models.
arXiv preprint arXiv:2307.09288
(2023)
\end{botherref}
\endbibitem

\bibitem[\protect\citeauthoryear{Yang et~al.}{2023}]{yang2023large}
\begin{botherref}
\oauthor{\bsnm{Yang}, \binits{R.}},
\oauthor{\bsnm{Tan}, \binits{T.F.}},
\oauthor{\bsnm{Lu}, \binits{W.}},
\oauthor{\bsnm{Thirunavukarasu}, \binits{A.J.}},
\oauthor{\bsnm{Ting}, \binits{D.S.W.}},
\oauthor{\bsnm{Liu}, \binits{N.}}:
Large language models in health care: Development, applications, and challenges.
Health Care Science
(2023)
\end{botherref}
\endbibitem

\bibitem[\protect\citeauthoryear{Li et~al.}{2022a}]{li2022ehrkit}
\begin{botherref}
\oauthor{\bsnm{Li}, \binits{I.}},
\oauthor{\bsnm{You}, \binits{K.}},
\oauthor{\bsnm{Tang}, \binits{X.}},
\oauthor{\bsnm{Qiao}, \binits{Y.}},
\oauthor{\bsnm{Huang}, \binits{L.}},
\oauthor{\bsnm{Hsieh}, \binits{C.-C.}},
\oauthor{\bsnm{Rosand}, \binits{B.}},
\oauthor{\bsnm{Goldwasser}, \binits{J.}},
\oauthor{\bsnm{Radev}, \binits{D.}}:
Ehrkit: A python natural language processing toolkit for electronic health record texts.
arXiv preprint arXiv:2204.06604
(2022)
\end{botherref}
\endbibitem

\bibitem[\protect\citeauthoryear{Li et~al.}{2022b}]{li2022neural}
\begin{barticle}
\bauthor{\bsnm{Li}, \binits{I.}},
\bauthor{\bsnm{Pan}, \binits{J.}},
\bauthor{\bsnm{Goldwasser}, \binits{J.}},
\bauthor{\bsnm{Verma}, \binits{N.}},
\bauthor{\bsnm{Wong}, \binits{W.P.}},
\bauthor{\bsnm{Nuzumlal{\i}}, \binits{M.Y.}},
\bauthor{\bsnm{Rosand}, \binits{B.}},
\bauthor{\bsnm{Li}, \binits{Y.}},
\bauthor{\bsnm{Zhang}, \binits{M.}},
\bauthor{\bsnm{Chang}, \binits{D.}}, \betal:
\batitle{Neural natural language processing for unstructured data in electronic health records: A review}.
\bjtitle{Computer Science Review}
\bvolume{46},
\bfpage{100511}
(\byear{2022})
\end{barticle}
\endbibitem

\bibitem[\protect\citeauthoryear{Xie et~al.}{2023}]{xie2023faithful}
\begin{botherref}
\oauthor{\bsnm{Xie}, \binits{Q.}},
\oauthor{\bsnm{Schenck}, \binits{E.J.}},
\oauthor{\bsnm{Yang}, \binits{H.S.}},
\oauthor{\bsnm{Chen}, \binits{Y.}},
\oauthor{\bsnm{Peng}, \binits{Y.}},
\oauthor{\bsnm{Wang}, \binits{F.}}:
Faithful ai in medicine: A systematic review with large language models and beyond.
Medrxiv: the Preprint Server for Health Sciences
(2023)
\end{botherref}
\endbibitem

\bibitem[\protect\citeauthoryear{Singhal et~al.}{2023}]{singhal2023expertlevel}
\begin{botherref}
\oauthor{\bsnm{Singhal}, \binits{K.}},
\oauthor{\bsnm{Tu}, \binits{T.}},
\oauthor{\bsnm{Gottweis}, \binits{J.}},
\oauthor{\bsnm{Sayres}, \binits{R.}},
\oauthor{\bsnm{Wulczyn}, \binits{E.}},
\oauthor{\bsnm{Hou}, \binits{L.}},
\oauthor{\bsnm{Clark}, \binits{K.}},
\oauthor{\bsnm{Pfohl}, \binits{S.}},
\oauthor{\bsnm{Cole-Lewis}, \binits{H.}},
\oauthor{\bsnm{Neal}, \binits{D.}},
\oauthor{\bsnm{Schaekermann}, \binits{M.}},
\oauthor{\bsnm{Wang}, \binits{A.}},
\oauthor{\bsnm{Amin}, \binits{M.}},
\oauthor{\bsnm{Lachgar}, \binits{S.}},
\oauthor{\bsnm{Mansfield}, \binits{P.}},
\oauthor{\bsnm{Prakash}, \binits{S.}},
\oauthor{\bsnm{Green}, \binits{B.}},
\oauthor{\bsnm{Dominowska}, \binits{E.}},
\oauthor{\bsnm{Arcas}, \binits{B.A.}},
\oauthor{\bsnm{Tomasev}, \binits{N.}},
\oauthor{\bsnm{Liu}, \binits{Y.}},
\oauthor{\bsnm{Wong}, \binits{R.}},
\oauthor{\bsnm{Semturs}, \binits{C.}},
\oauthor{\bsnm{Mahdavi}, \binits{S.S.}},
\oauthor{\bsnm{Barral}, \binits{J.}},
\oauthor{\bsnm{Webster}, \binits{D.}},
\oauthor{\bsnm{Corrado}, \binits{G.S.}},
\oauthor{\bsnm{Matias}, \binits{Y.}},
\oauthor{\bsnm{Azizi}, \binits{S.}},
\oauthor{\bsnm{Karthikesalingam}, \binits{A.}},
\oauthor{\bsnm{Natarajan}, \binits{V.}}:
Towards Expert-Level Medical Question Answering with Large Language Models
(2023)
\end{botherref}
\endbibitem

\bibitem[\protect\citeauthoryear{Xu et~al.}{2023}]{xu2023baize}
\begin{botherref}
\oauthor{\bsnm{Xu}, \binits{C.}},
\oauthor{\bsnm{Guo}, \binits{D.}},
\oauthor{\bsnm{Duan}, \binits{N.}},
\oauthor{\bsnm{McAuley}, \binits{J.}}:
Baize: An open-source chat model with parameter-efficient tuning on self-chat data.
arXiv preprint arXiv:2304.01196
(2023)
\end{botherref}
\endbibitem

\bibitem[\protect\citeauthoryear{Hiesinger et~al.}{2023}]{hiesinger2023almanac}
\begin{botherref}
\oauthor{\bsnm{Hiesinger}, \binits{W.}},
\oauthor{\bsnm{Zakka}, \binits{C.}},
\oauthor{\bsnm{Chaurasia}, \binits{A.}},
\oauthor{\bsnm{Shad}, \binits{R.}},
\oauthor{\bsnm{Dalal}, \binits{A.}},
\oauthor{\bsnm{Kim}, \binits{J.}},
\oauthor{\bsnm{Moor}, \binits{M.}},
\oauthor{\bsnm{Alexander}, \binits{K.}},
\oauthor{\bsnm{Ashley}, \binits{E.}},
\oauthor{\bsnm{Boyd}, \binits{J.}}, et al.:
Almanac: Retrieval-augmented language models for clinical medicine
(2023)
\end{botherref}
\endbibitem

\bibitem[\protect\citeauthoryear{Long et~al.}{2023}]{long2023chatent}
\begin{botherref}
\oauthor{\bsnm{Long}, \binits{C.}},
\oauthor{\bsnm{Subburam}, \binits{D.}},
\oauthor{\bsnm{Lowe}, \binits{K.}},
\oauthor{\bsnm{Santos}, \binits{A.d.}},
\oauthor{\bsnm{Zhang}, \binits{J.}},
\oauthor{\bsnm{Hwang}, \binits{S.}},
\oauthor{\bsnm{Saduka}, \binits{N.}},
\oauthor{\bsnm{Horev}, \binits{Y.}},
\oauthor{\bsnm{Su}, \binits{T.}},
\oauthor{\bsnm{Cote}, \binits{D.}}, et al.:
Chatent: Augmented large language model for expert knowledge retrieval in otolaryngology-head and neck surgery.
medRxiv,
2023--08
(2023)
\end{botherref}
\endbibitem

\bibitem[\protect\citeauthoryear{Malaviya et~al.}{2023}]{malaviya2023expertqa}
\begin{botherref}
\oauthor{\bsnm{Malaviya}, \binits{C.}},
\oauthor{\bsnm{Lee}, \binits{S.}},
\oauthor{\bsnm{Chen}, \binits{S.}},
\oauthor{\bsnm{Sieber}, \binits{E.}},
\oauthor{\bsnm{Yatskar}, \binits{M.}},
\oauthor{\bsnm{Roth}, \binits{D.}}:
ExpertQA: Expert-Curated Questions and Attributed Answers
(2023)
\end{botherref}
\endbibitem

\bibitem[\protect\citeauthoryear{Bodenreider}{2004}]{bodenreider2004unified}
\begin{barticle}
\bauthor{\bsnm{Bodenreider}, \binits{O.}}:
\batitle{The unified medical language system (umls): integrating biomedical terminology}.
\bjtitle{Nucleic acids research}
\bvolume{32}(\bissue{suppl\_1}),
\bfpage{267}--\blpage{270}
(\byear{2004})
\end{barticle}
\endbibitem

\bibitem[\protect\citeauthoryear{Huang et~al.}{2020}]{huang2020biomedical}
\begin{botherref}
\oauthor{\bsnm{Huang}, \binits{K.-H.}},
\oauthor{\bsnm{Yang}, \binits{M.}},
\oauthor{\bsnm{Peng}, \binits{N.}}:
Biomedical event extraction with hierarchical knowledge graphs.
arXiv preprint arXiv:2009.09335
(2020)
\end{botherref}
\endbibitem

\bibitem[\protect\citeauthoryear{He et~al.}{2019}]{he2019integrating}
\begin{botherref}
\oauthor{\bsnm{He}, \binits{B.}},
\oauthor{\bsnm{Zhou}, \binits{D.}},
\oauthor{\bsnm{Xiao}, \binits{J.}},
\oauthor{\bsnm{Liu}, \binits{Q.}},
\oauthor{\bsnm{Yuan}, \binits{N.J.}},
\oauthor{\bsnm{Xu}, \binits{T.}}, et al.:
Integrating graph contextualized knowledge into pre-trained language models.
arXiv preprint arXiv:1912.00147
(2019)
\end{botherref}
\endbibitem

\bibitem[\protect\citeauthoryear{Lu et~al.}{2021}]{lu2021parameter}
\begin{bchapter}
\bauthor{\bsnm{Lu}, \binits{Q.}},
\bauthor{\bsnm{Dou}, \binits{D.}},
\bauthor{\bsnm{Nguyen}, \binits{T.H.}}:
\bctitle{Parameter-efficient domain knowledge integration from multiple sources for biomedical pre-trained language models}.
In: \bbtitle{Findings of the Association for Computational Linguistics: EMNLP 2021},
pp. \bfpage{3855}--\blpage{3865}
(\byear{2021})
\end{bchapter}
\endbibitem

\bibitem[\protect\citeauthoryear{Aracena et~al.}{2022}]{aracena2022knowledge}
\begin{bchapter}
\bauthor{\bsnm{Aracena}, \binits{C.}},
\bauthor{\bsnm{Villena}, \binits{F.}},
\bauthor{\bsnm{Rojas}, \binits{M.}},
\bauthor{\bsnm{Dunstan}, \binits{J.}}:
\bctitle{A knowledge-graph-based intrinsic test for benchmarking medical concept embeddings and pretrained language models}.
In: \bbtitle{Proceedings of the 13th International Workshop on Health Text Mining and Information Analysis (LOUHI)},
pp. \bfpage{197}--\blpage{206}
(\byear{2022})
\end{bchapter}
\endbibitem

\bibitem[\protect\citeauthoryear{Gao et~al.}{2023}]{gao2023leveraging}
\begin{botherref}
\oauthor{\bsnm{Gao}, \binits{Y.}},
\oauthor{\bsnm{Li}, \binits{R.}},
\oauthor{\bsnm{Caskey}, \binits{J.}},
\oauthor{\bsnm{Dligach}, \binits{D.}},
\oauthor{\bsnm{Miller}, \binits{T.}},
\oauthor{\bsnm{Churpek}, \binits{M.M.}},
\oauthor{\bsnm{Afshar}, \binits{M.}}:
Leveraging a medical knowledge graph into large language models for diagnosis prediction.
arXiv preprint arXiv:2308.14321
(2023)
\end{botherref}
\endbibitem

\bibitem[\protect\citeauthoryear{Raza et~al.}{2022}]{raza2022large}
\begin{barticle}
\bauthor{\bsnm{Raza}, \binits{S.}},
\bauthor{\bsnm{Reji}, \binits{D.J.}},
\bauthor{\bsnm{Shajan}, \binits{F.}},
\bauthor{\bsnm{Bashir}, \binits{S.R.}}:
\batitle{Large-scale application of named entity recognition to biomedicine and epidemiology}.
\bjtitle{PLOS Digital Health}
\bvolume{1}(\bissue{12}),
\bfpage{0000152}
(\byear{2022})
\end{barticle}
\endbibitem

\bibitem[\protect\citeauthoryear{Li et~al.}{2021}]{li2021unsupervised}
\begin{botherref}
\oauthor{\bsnm{Li}, \binits{I.}},
\oauthor{\bsnm{Yan}, \binits{V.}},
\oauthor{\bsnm{Li}, \binits{T.}},
\oauthor{\bsnm{Qu}, \binits{R.}},
\oauthor{\bsnm{Radev}, \binits{D.}}:
Unsupervised cross-domain prerequisite chain learning using variational graph autoencoders.
arXiv preprint arXiv:2105.03505
(2021)
\end{botherref}
\endbibitem

\bibitem[\protect\citeauthoryear{Abacha et~al.}{2017}]{abacha2017overview}
\begin{bchapter}
\bauthor{\bsnm{Abacha}, \binits{A.B.}},
\bauthor{\bsnm{Agichtein}, \binits{E.}},
\bauthor{\bsnm{Pinter}, \binits{Y.}},
\bauthor{\bsnm{Demner-Fushman}, \binits{D.}}:
\bctitle{Overview of the medical question answering task at trec 2017 liveqa.}
In: \bbtitle{TREC},
pp. \bfpage{1}--\blpage{12}
(\byear{2017})
\end{bchapter}
\endbibitem

\bibitem[\protect\citeauthoryear{Lin}{2004}]{lin2004rouge}
\begin{bchapter}
\bauthor{\bsnm{Lin}, \binits{C.-Y.}}:
\bctitle{Rouge: A package for automatic evaluation of summaries}.
In: \bbtitle{Text Summarization Branches Out},
pp. \bfpage{74}--\blpage{81}
(\byear{2004})
\end{bchapter}
\endbibitem

\bibitem[\protect\citeauthoryear{Zhang et~al.}{2019}]{zhang2019bertscore}
\begin{botherref}
\oauthor{\bsnm{Zhang}, \binits{T.}},
\oauthor{\bsnm{Kishore}, \binits{V.}},
\oauthor{\bsnm{Wu}, \binits{F.}},
\oauthor{\bsnm{Weinberger}, \binits{K.Q.}},
\oauthor{\bsnm{Artzi}, \binits{Y.}}:
Bertscore: Evaluating text generation with bert.
arXiv preprint arXiv:1904.09675
(2019)
\end{botherref}
\endbibitem

\end{thebibliography}
\clearpage
\appendix
\section*{Appendix}
\section*{Case Study \\(ChatGPT-3.5 \& UMLS-augmented ChatGPT-3.5)}

\begin{table}[h]
    \centering
    \begin{tabular}{p{\linewidth}}
        \toprule
        \textbf{Question:} Do 5 mg. Zolmitriptan tabzets contain gluten?\\
        \midrule
        \textbf{Model:} ChatGPT-3.5 \\
        \midrule
        \textcolor{red}{I'm sorry, but as an AI language model, I don't have access to specific drug information or the ability to browse the internet.} To determine if a specific medication contains gluten, it is best to consult the medication's packaging, the manufacturer's website, or consult a healthcare professional or pharmacist who can provide accurate and up-to-date information. \\
        \midrule
        \textbf{Model:} UMLS-augmented ChatGPT-3.5 \\
        \midrule
        \textcolor{red}{Yes, 5 mg Zolmitriptan tablets do not contain gluten.} Zolmitriptan is a member of the triptan class of agents with anti-migraine properties. It selectively binds to and activates serotonin (5-HT) 1B receptors expressed in intracranial arteries and 5-HT 1D receptors located on peripheral trigeminal sensory nerve terminals in the meninges and central terminals in brainstem sensory nuclei. This receptor binding results in the constriction of cranial vessels, reduction of vessel pulsation, and inhibition of nociceptive transmission, providing relief of migraine headaches.\\
        Regarding gluten, it is important to note that gluten is a protein found in wheat, barley, and rye. It is commonly avoided by individuals with celiac disease or gluten sensitivity. However, Zolmitriptan tablets do not contain gluten, so they can be safely used by individuals who need to avoid gluten.\\
        If you have any further questions or concerns, please let me know.\\
    \bottomrule
    \end{tabular}
\end{table}

\begin{table}[h]
    \centering
    \begin{tabular}{p{\linewidth}}
        \toprule
        \textbf{Question:} Does fertilization of an egg have to take place before there can be a molar pregnancy? If so, when does it take place?\\
        \midrule
        \textbf{Model:} ChatGPT-3.5 \\
        \midrule
        \textcolor{red}{No, fertilization of an egg does not have to take place before a molar pregnancy can occur.} In fact, a molar pregnancy is a type of abnormal pregnancy where fertilization does occur, but the resulting embryo develops abnormally.\\
        In a molar pregnancy, instead of a normal embryo developing, there is an overgrowth of placental tissue. This occurs when there is an error during fertilization, leading to an abnormal number of chromosomes in the fertilized egg. This abnormal fertilization can happen either when an egg is fertilized by two sperm (complete molar pregnancy) or when a normal egg is fertilized by a single sperm but duplicates its own chromosomes (partial molar pregnancy).\\
        So, fertilization does take place in a molar pregnancy, but the resulting pregnancy is abnormal due to chromosomal errors.\\
        \midrule
        \textbf{Model:} UMLS-augmented ChatGPT-3.5 \\
        \midrule
        \textcolor{red}{Yes, fertilization of an egg must take place before there can be a molar pregnancy.} Fertilization is the process in which a spermatozoon (sperm) fuses with an ovum (egg) to form a zygote. This fusion of genetic material is necessary for the development of a pregnancy.\\
        In the case of a molar pregnancy, there is an abnormal growth of trophoblastic tissue, which is the tissue that normally surrounds and nourishes the embryo. This abnormal growth can occur when there is an error during fertilization, leading to the development of a hydatidiform mole.\\
        A hydatidiform mole is characterized by the swelling of the chorionic villi, which are finger-like projections in the placenta. This swelling is accompanied by elevated levels of human chorionic gonadotropin (hCG), a hormone produced during pregnancy. Hydatidiform moles can be categorized as complete or partial based on their gross morphology, histopathology, and karyotype.\\
        To summarize, fertilization of an egg is necessary for the development of a molar pregnancy. If there is an error during fertilization, it can lead to the formation of a hydatidiform mole.\\
    \bottomrule
    \end{tabular}
\end{table}

\end{document}